\pdfoutput=1
\documentclass[letterpaper]{article}
\usepackage[preprint]{aaai2027}
\usepackage[hyphens]{url}
\usepackage{graphicx}
\usepackage{natbib}
\usepackage{caption}
\usepackage{amsmath,amssymb}
\usepackage{multirow}
\usepackage{bm}
\usepackage{hhline}
\usepackage[table]{xcolor}
\definecolor{highlight}{RGB}{230,242,255}

\usepackage{booktabs}

\title{DOW-KE: Anchor-Free Multi-Layer Knowledge Editing via Direct End-to-End Weight Optimization}
\author{Ran Chen,
	Junbo Zhang,
	Qianli Zhou,
	Xinyang Deng,
	Wen Jiang\corresponding}
\affiliations{School of Electronics and Information, Northwestern Polytechnical University}

\begin{document}
	
\maketitle

\begin{abstract}
	Multi-layer locate-then-edit methods for knowledge editing first optimize target residual-stream activations (anchors) at selected layers, then realize them layer by layer as weight updates. This pipeline optimizes an intermediate representation but deploys multi-layer weight updates whose joint effect through the true forward pass is never itself optimized: regardless of how anchors are set or propagated, each update comes from a local solve, so propagation-induced attenuation and distortion go uncorrected, leaving a closure gap between anchor targets and realized edits. We propose DOW-KE, an anchor-free method built on a single principle: what is optimized must be exactly what is deployed. DOW-KE backpropagates the final editing objective through the complete model, jointly optimizing the updates of all edited layers so cross-layer propagation and coupling enter every gradient step. The same principle dictates where preservation resides: embedding the preservation projection in the update parameterization, inside the computation graph, makes every gradient act on the deployed update; post-hoc constraints would reopen the gap, and the constrained search keeps edits clear of protected knowledge. In large-scale sequential editing on two datasets and three models, DOW-KE achieves the highest overall Score and neighborhood Specificity in five of six model–dataset settings among the evaluated baselines.
\end{abstract}

\section{Introduction}

Large language models (LLMs) store extensive cross-domain knowledge in their parameters, enabled by their massive scale and broad training corpora. Yet this knowledge is encoded implicitly and distributed across parameters rather than stored as directly addressable entries \cite{geva2023dissecting}, making individual facts difficult to modify \cite{mitchell2022mend,lewis2020retrieval,hase2023localization}. Corpus-level knowledge updates are commonly performed through continued pretraining or fine-tuning \cite{jiang2024instruction}, but their substantial time and computational costs and limited target specificity \cite{decao2021editing,mitchell2022mend} make them uneconomical for requests to modify a specific item. Model editing is designed for precisely this setting: it makes precise and localized changes to specified knowledge in model parameters at minimal intervention cost \cite{sinitsin2020editable,yao2023editing}.
\begin{figure}[t]
	\centering
	\includegraphics[width=1.0\columnwidth]{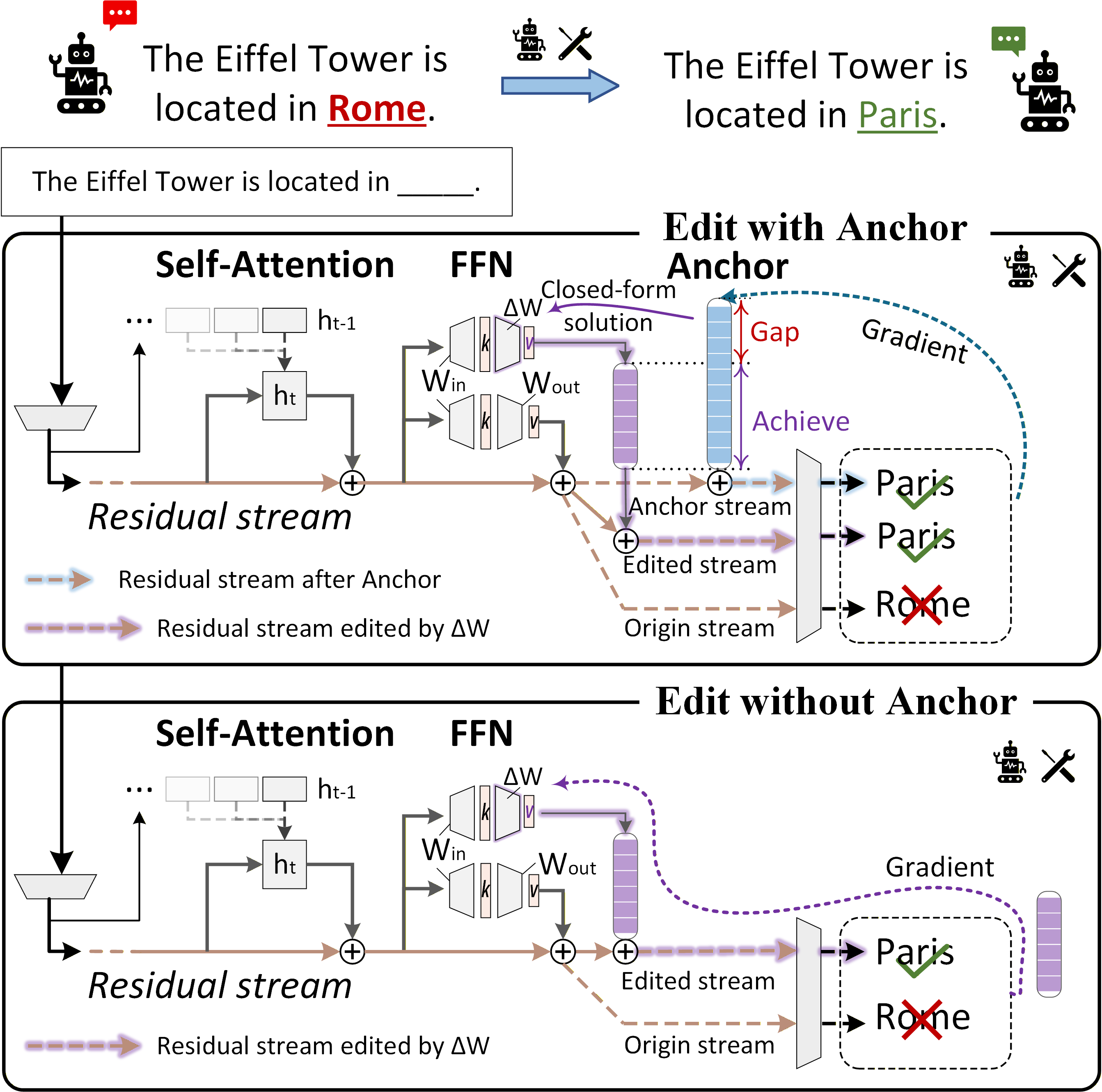} 
	\caption{Schematic illustration of knowledge editing.  Knowledge editing enables precise modification of model knowledge by updating only the output weights of the feed-forward layers. Most existing methods rely on anchors to route optimization signals, whereas our method eliminates this anchor-based structure and directly optimizes the weights using gradients.}
	\label{graphic_abstract}
\end{figure}

Locate-then-edit is a mainstream parameter-editing paradigm: it identifies a small subset of parameters associated with the target knowledge and directly rewrites them without introducing auxiliary modules. ROME is a representative single-layer method \cite{meng2022rome}. After freezing the weights, it optimizes a replacement activation at the selected layer's feed-forward-network output for the subject's final token, so that the model produces the edited answer for both the edit prompt and its prefix variants. We refer to this layer–token position as the \textit{anchor} and the resulting activation as the anchor target. ROME then computes a closed-form least-squares update to the layer using the anchor target as its sole constraint. Because a single-layer edit requires one linear operator to balance efficacy and locality, its capacity is limited. Multi-layer methods therefore retain a single anchor but distribute the update across preceding shallower layers, assuming that the layerwise increments propagate approximately losslessly through the residual stream and combine linearly into the target residual at the anchor \cite{meng2023memit}.

Although multi-layer editing expands editing capacity, its implicit linear-superposition assumption does not hold, leaving a gap between the edit increment delivered to the anchor and the intended target. Prior work has attempted to address this problem \cite{li2025rethinking,liu2026forwardreplay}, but these methods overlook the central issue: regardless of where the anchor is placed or how it is propagated, each layerwise weight update is still obtained by locally solving for a predetermined intermediate target, while the terminal effect produced by all layerwise updates is never optimized at any stage.

This issue is not a harmless abstraction; it produces two linked adverse effects within the two-stage framework. First, propagation costs remain invisible. As the increments written at different layers travel through the residual stream, they attenuate, become distorted, and interact with subsequent computations \cite{liu2026forwardreplay}. Because no stage includes the terminal effect in its objective, these losses can neither be detected nor corrected. Second, when delivery is unobserved, write strength can only be set blindly. Existing methods compensate by building in a margin: they optimize the target activation beyond the necessary strength and then repeatedly recompute the residual layer by layer to absorb the gap. This process is one-way, and a layer cannot be revisited once it has been written. Without terminal feedback, the compensation margin cannot be calibrated, creating an unavoidable trade-off under any single strength setting: insufficient strength yields incomplete edits, whereas excessive strength causes unnecessary updates to spill into unrelated knowledge \cite{hartvigsen2023grace}.

\begin{figure}[t]
	\centering
	\includegraphics[width=0.96\columnwidth]{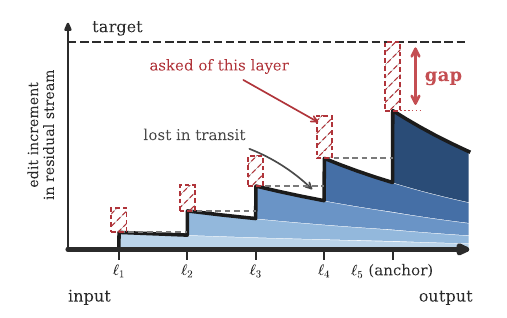}
	\caption{The edit increment never reaches its target. Layers are solved shallow-to-deep: each is asked for a share of the shortfall then measured at the anchor, writes only part of it (hatched blocks), and what it writes decays in transit (grey levels held from each peak). Both losses accumulate into the gap still open at the anchor.}
	\label{fig:teaser}
\end{figure}
These two effects share the same root cause: the object being optimized differs from the object ultimately deployed, a discrepancy we term structural misalignment. DOW-KE therefore follows a single design principle: make the two identical. Rather than prescribing intermediate targets, it uses the final editing objective as the loss and jointly optimizes the weight updates at all edited layers through full-model backpropagation. By removing the anchor-based gradient cutoff, DOW-KE allows attenuation, distortion, cross-layer coupling, and subspace constraints to be reflected directly in the gradients, eliminating the need for a delivery assumption. Optimization itself coordinates the division of labor across layers instead of following a predefined schedule. Write strength is learned jointly across layers rather than preset by residual allocation.

Our contributions are summarized as follows:

\begin{itemize}
\item  We identify and mechanistically characterize a long-standing structural misalignment in two-stage, multi-layer locate-then-edit methods, and quantitatively validate it through controlled experiments.
\item  We propose DOW-KE, an anchor-free multi-layer editing method that uses the final editing objective as its loss and jointly optimizes the weight updates at all edited layers through full-model backpropagation, thereby explicitly accounting for propagation attenuation, distortion, and cross-layer coupling. It also embeds all preservation constraints into the update parameterization.
\item In sequential editing experiments on two datasets and three models, DOW-KE achieves the highest overall Score in five of six model-dataset settings among the evaluated baselines. Ablations further characterize the behavior of direct weight optimization and in-graph constraint parameterization in the evaluated setting.
\end{itemize}

\section{Related Work}

\subsection{Locate-then-Edit Methods}

ROME~\cite{meng2022rome} pioneered this paradigm: causal tracing locates the layers carrying a fact, and a two-stage procedure optimizes an anchor target by gradient descent before converting it into a closed-form weight update. MEMIT~\cite{meng2023memit} distributes the residual across several layers to increase capacity; AlphaEdit~\cite{fang2025alphaedit} constrains the update to the null space of protected-knowledge keys, and EvoEdit~\cite{lyu2026evoedit} extends that projection to cover prior edits. The attenuation and distortion introduced by pairing multi-layer editing with a single anchor, however, went largely unexamined until BLUE \cite{li2025rethinking} and FE \cite{liu2026forwardreplay}: the former introduces an additional anchor, assigning one anchor target to each boundary layer, at a cost growing with their number; the latter constructs layerwise targets by forward replay. Progress along this line has thus refined the intermediate targets and the coverage of preservation statistics, while leaving the two-stage premise itself intact: what is optimized is an activation at the anchor, not the weight update actually deployed.

\subsection{Other Knowledge-Editing Paradigms}

Beyond methods that directly modify the original parameters, knowledge editing encompasses two other paradigms. The first uses auxiliary parameters or external memory: the original model remains unchanged, while new knowledge is stored in an additional component. 
WISE and MELO store edits in routed side memories or dynamic LoRA modules \cite{wang2024wise,yu2024melo}. The second paradigm is meta-learning, which trains an auxiliary network to learn how to modify parameters. KE and MEND use hypernetworks to transform naive fine-tuning gradients into parameter updates that satisfy locality constraints \cite{decao2021editing,mitchell2022mend}.

External-memory methods are particularly suited to lifelong editing, where edits arrive one by one and must take effect immediately; however, their storage and routing costs grow continually with the number of edits. LocFT-BF revisits direct fine-tuning by combining localized parameter selection with breadth-first optimization \cite{yang2026finetuning}.  Under the same setting considered by WISE \cite{wang2024wise}, the concurrent work LOKI (June 2026; \citealp{eskandar2026loki} ) instead directly modifies the model's internal weights. LOKI removes the intermediate activation anchor. Its weight-space null-space constraint specifies only a general feasible update space and does not use the target edit to predetermine which input features the update should respond to. Consequently, full-size weight optimization jointly determines which input features to affect and what content to write. By contrast, DOW-KE retains the batched locate-then-edit mechanism, in which edit keys and preservation statistics predetermine which input features an update can respond to, and learns only what to write end to end against the final editing objective. This division keeps each update associated with a specific edit fact, restricts optimization to a low-rank space that grows with the edit batch, and supports joint coordination among edits within the batch. DOW-KE therefore remains a fact-directed locate-then-edit method, replacing the anchor target with the final editing objective; LOKI instead learns the update structure within its general feasible weight space after removing the intermediate activation anchor.

\section{Preliminaries}
\subsection{Residual Stream and FFN Key-Value Memory}
Given an autoregressive language model $G$, let $\mathbf{h}_t^{l}\in\mathbb{R}^{d_1}$ denote the residual-stream state at position $t$ in layer $l$. Let $\mathbf{a}_t^{l}$ and $\mathbf{m}_t^{l}$ denote the increments written to the residual stream by the attention module and the FFN, respectively. We use $\gamma$ to denote layer normalization and $\phi^l$ to denote the FFN feature map before the down projection. The down-projection matrix to be edited is denoted by $W^{l}\in\mathbf{R}^{d_1\times d_0}$. We define the input and output of the down-projection matrix as the key and value:

\begin{equation}
\mathbf{k}_t^{l}\triangleq\phi^l\!\big(\gamma(\mathbf{h}_t^{l-1}+\mathbf{a}_t^{l})\big)\in\mathbf{R}^{d_0},
\qquad
\mathbf{v}_t^{l}\triangleq W^{l}\mathbf{k}_t^{l}\in\mathbb{R}^{d_1}.
\label{eq:key and value}
\end{equation}

The feature map can be written in simplified form as $\phi^l(\mathbf{x})=\sigma(W_{\mathrm{in}}^l\mathbf{x})$. Under the linear associative-memory view \cite{meng2022rome}, $W^{l}$ stores mappings from input patterns to residual-stream increments, i.e., $W^{l}\mathbf{k}=\mathbf{v}$. Locate-then-edit methods accordingly formulate knowledge updates as rewriting specific key-value associations.




\subsection{Anchor-Target Optimization}
After localization, ROME, MEMIT, and AlphaEdit first optimize a target activation at the deepest edited layer $l_L$. Let $G_{i,\boldsymbol{\delta}}$ denote the model obtained by adding an increment $\boldsymbol{\delta}$ to the state $\mathbf{h}_i^{l_L}$ at the subject's final token for the $i$-th request. Let $\mathcal{L}_{\mathrm{edit}}$ denote the negative log-likelihood of the target object, and let $\mathcal{L}_{\mathrm{KL}}$ constrain the output distributions before and after the intervention:
\begin{equation}
\boldsymbol{\delta}_i=\arg\min_{\boldsymbol{\delta}}\;
\mathcal{L}_{\mathrm{edit}}(G_{i,\boldsymbol{\delta}})
+\lambda_{\mathrm{KL}}\mathcal{L}_{\mathrm{KL}}(G_{i,\boldsymbol{\delta}},G).
\label{eq:anchor loss}
\end{equation}
$\mathcal{L}_{\mathrm{edit}}$ is averaged over $N$ random prefixes $x_j$ to make the target activation robust to contextual variation. $\mathcal{L}_{\mathrm{KL}}$ is evaluated on the essence prompt $p_i'$ to suppress global drift in the subject representation. The optimized target state is
\begin{equation}
\mathbf{z}_i=\mathbf{h}_i^{l_L}+\boldsymbol{\delta}_i.
\label{eq:z_computer}
\end{equation}
Consistent with the terminology in the Introduction, this state is the anchor target. It specifies the residual-stream state that the model should reach at the deepest edited layer but does not directly modify the weights. To avoid confusion with the FFN output value $\mathbf{v}$, we follow MEMIT and denote this state by $\mathbf{z}$.

\subsection{Closed-Form Weight Update}

The second stage converts the anchor target into weight updates at one or more edited layers. For any layer $l$ to be updated, the key for each request is first averaged over random prefixes. The resulting vectors form the edit-key matrix $K_1=[\mathbf{k}_1^{l},\ldots,\mathbf{k}_u^{l}]\in\mathbf{R}^{d_0\times u}$ for a batch of $u$ requests. Let $\mathbf{r}_i$ denote the portion of the anchor-target residual assigned to the current layer for request $i$. Its target value is then $\mathbf{v}_{1,i}=W\mathbf{k}_i^{l}+\mathbf{r}_i$. When the entire residual is assigned to the deepest edited layer, $\mathbf{r}_i=\mathbf{z}_i-\mathbf{h}_i^{l_L}$. Collecting all target values into $V_1$, the weight update $\Delta$ is obtained from the following least-squares problem, in which $K_0$ and $V_0=WK_0$ are the keys and values of knowledge that must be preserved:
\begin{equation}
	\Delta^{\star}=\arg\min_{\Delta}
	\big\|(W+\Delta)K_1-V_1\big\|_F^{2}
	+\big\|(W+\Delta)K_0-V_0\big\|_F^{2}.
	\label{eq:delta L2 norm}
\end{equation}
Let $R\triangleq V_1-WK_1$ denote the value residual. Using $WK_0=V_0$, the problem above reduces to regularized least squares in which $R$ specifies the desired correction, while the second-order statistics of the edit and preserved keys jointly constrain the solution directions. In practice, $K_0K_0^{\top}$ is approximated by $\lambda C$, with $C$ the second moment of keys from a general-domain corpus, yielding
\begin{equation}
	\Delta^{\star}=R K_1^{\top}
	\big(\lambda C+K_1K_1^{\top}\big)^{-1}.
	\label{eq:closed-form delta W}
\end{equation}
This form handles a batch of requests simultaneously and serves as the basis for the per-layer closed-form weight update in multi-layer editing.

MEMIT distributes the same batch of requests across a layer window $\mathcal{L}$ to alleviate the capacity limitation of a single layer. The anchor target is optimized only once at the deepest layer, after which the layerwise updates are computed sequentially from shallow to deep. When processing the $j$-th edited layer $l_j$, the updates to the preceding $j-1$ layers have already taken effect. The algorithm performs a new forward pass to recompute the current $\mathbf{h}_i^{l_L}$ and edit keys $K_1^{l_j}$, and then evenly distributes the remaining anchor-target residual across the remaining layers:
\begin{equation}
	\mathbf{r}_i^{(j)}=
	\frac{\mathbf{z}_i-\mathbf{h}_i^{l_L}}{L-j+1},
	\qquad
	R^{(j)}=\big[\mathbf{r}_1^{(j)},\ldots,\mathbf{r}_u^{(j)}\big].
	\label{eq:multi-layer assign}
\end{equation}
The update to the current layer is
\begin{equation}
	\Delta^{l_j}=R^{(j)}\big(K_1^{l_j}\big)^{\top}
	\Big(\lambda C^{l_j}+K_1^{l_j}\big(K_1^{l_j}\big)^{\top}\Big)^{-1}.
	\label{eq:multi-layer closed-form delta W}
\end{equation}
Each layerwise closed-form solution fits only the local target assigned to that layer and does not directly enforce attainment of the final anchor target. 
\section{Empirical Study}
\subsection{Single-Anchor Multi-Layer Editing}
\begin{figure*}[t]
	\centering
	\includegraphics[width=\textwidth]{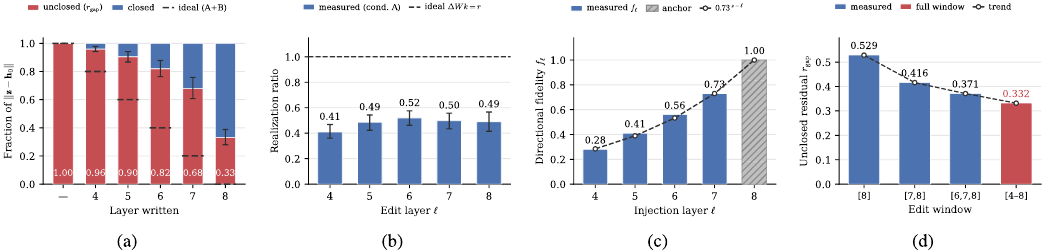}
	\caption{Two-stage mediation leaves the anchor target unmet
		(AlphaEdit, Llama-3-8B, CounterFact).
		(a) Residual closure as the edit window is written shallow-to-deep
		(x-axis: deepest layer written, ``\textemdash'' = pre-edit; red = still
		unclosed, blue = closed; the dashed tick is the ideal split under conditions
		A+B, which would reach $0$); the terminal red segment is
		$r_{\mathrm{gap}}\approx0.33$.
		(b) Per-layer realization ratio
		$\langle\Delta W k, r\rangle/\|r\|^2$ (condition A); ideal $=1$.
		(c) Forward directional fidelity $f_\ell$ (cosine between the injected
		target direction and the direction arriving at the anchor); layer~$8$ is the
		anchor (no transport); dashed guide is $0.73^{\,z-\ell}$.
		(d) Terminal $r_{\mathrm{gap}}$ vs.\ edit window (1/2/3/5 layers $=$
		[8]/[7,8]/[6,7,8]/[4--8]); the gap shrinks but never closes.}
	\label{fig:motivation}
\end{figure*}
\begin{figure}[t]
	\centering
	\includegraphics[width=\columnwidth]{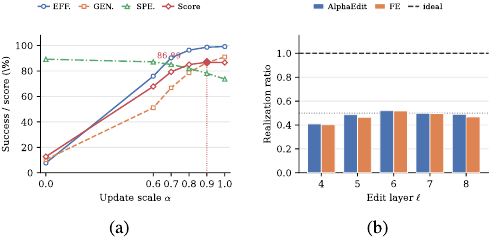}
	\caption{The gap resists cheap fixes.
		(a) Post-hoc update scale $\alpha$: efficacy (EFF), generalization
		(GEN), specificity (SPE), and the composite score; EFF/GEN and SPE move in
		opposite directions on one trade-off curve, and the composite peaks at
		$86.89$ ($\alpha{=}0.9$)---rescaling alone cannot close the gap.
		(b) Realization ratio, standard AlphaEdit (uniform apportionment)
		vs.\ FE (forward-replay targets); both stay near $0.5$ regardless of how the
		target is constructed, so better targets do not resolve the realization error.
		}
	\label{fig:remedy}
\end{figure}
To support batch editing, single-anchor multi-layer methods usually distribute updates across multiple layers while retaining a single anchor. The anchor target is still optimized only once at the deepest edited layer, while the residual $\mathbf{z}_i-\mathbf{h}_i^{l_L}$ is allocated across layers with the quota $1/(L-j+1)$, and each layer independently performs a least-squares weight update. This decomposition is no longer lossless. Its closure depends on two conditions that hold automatically in the single-layer setting: (A) each layer update exactly realizes its allocated target at the edit key, i.e., $\Delta^{l_j}\mathbf{k}_i^{l_j}=\mathbf{r}_i^{(j)}$; and (B) the increment reaches the deepest edited layer losslessly through the residual stream without interference from increments induced by other requests in the same batch. If both conditions hold, layerwise recursion yields exact closure at the deepest layer, $\mathbf{h}_i^{l_L}=\mathbf{z}_i$. 

Empirically, this is not the case. Figure \ref{fig:motivation}(a) shows the fraction of the residual that remains unrealized at each edited layer. Its terminal value,
$r_{\mathrm{gap}}=\big\|\mathbf{z}_i-\tilde{\mathbf{h}}_i^{l_L}\big\|\big/\big\|\mathbf{z}_i-\mathbf{h}_i^{l_L}\big\|\approx0.33$, jointly measures the extent to which the two conditions are violated. 

Condition (A) fails because the closed-form solution does not fit only the edit target. To preserve unrelated knowledge, it incorporates a preservation term in addition to the edit term—MEMIT's preservation term or AlphaEdit's null-space projection—and balances the two within the same least-squares problem. Target realization at the edit key is therefore systematically discounted. Figure~\ref{fig:motivation}(b) measures this discount locally at the updated layer, without transport effects.

Condition (B) fails during transport: subsequent layers dilute the increment's relative contribution, while unplanned attention and FFN responses distort its direction, as shown in Figure \ref{fig:motivation}(c). 
The remaining residual is recomputed before every layer update, so any deficit at a shallow layer is folded into the targets of subsequent layers. Nevertheless, because realization at every layer is itself lossy, the net result along the chain is an unrealized residual at the terminal layer. The implication is that the property verified at the anchor target is never actually deployed: the edited model ends at an intermediate state that was not optimized at any stage.

Figure \ref{fig:motivation}(d) shows that the unrealized residual decreases monotonically as the number of edited layers increases, consistent with the superior performance of multi-layer editing. Multi-layer editing enables multiple lossy realizations of the same target to be performed effectively; however, a gap persists even when the full layer window is used. A direct remedy would be to scale all increments globally to fill this gap. Scaling, however, does not alter the update structure; it proportionally amplifies the update's effect at every position and on every unrelated key. 
As shown by the scaling-coefficient sweep in Figure~\ref{fig:remedy}(a), editing efficacy and locality move in opposite directions along the same trade-off curve, and no scaling factor improves both simultaneously. The gap cannot be eliminated by changing amplitude alone; doing so requires changing the optimization structure.

\subsection{Multiple Anchors and Multi-Layer Editing}

Several studies have identified the issue above and proposed two successive remedies at the level of target specification. BLUE edits only the first and last boundary layers of the editing window and introduces a native anchor target at each layer. Each layer independently performs full gradient-based anchor-target optimization instead of receiving an allocated residual; because each target is optimized through the actual forward computation, downstream transport is already taken into account. This design incurs two costs. First, gradient-based anchor-target optimization is the most expensive stage of the pipeline and doubling the number of anchors doubles this cost. Second, the layerwise anchor targets are optimized independently, with no mechanism to guarantee their cross-layer compatibility. FE further argues that independently specified layerwise targets are mutually incompatible. It therefore optimizes an anchor target only at the shallowest edited layer and then uses a forward pass to convert it at low cost into mutually compatible targets for subsequent layers.

However, both approaches improve only the targets. Each layer update is still realized by the same closed-form solution, while realization errors have no feedback channel: once introduced, they propagate and accumulate through the residual stream. FE also explicitly acknowledges that the layerwise increments are not guaranteed to faithfully realize the targets specified by the anchors. Our measurements in Figure \ref{fig:remedy}(b) confirm this point: the layerwise realization errors under FE-style targets nearly overlap those under standard residual-allocation targets. Realization loss is largely independent of how the targets are specified; it is a property of the closed-form weight update itself. As long as the two-stage structure of “specify a target, then realize it” remains unchanged, better targets still cannot be guaranteed to be attained.

\subsection{Removing the Intermediate Anchor}
Both remedies stay inside the same information boundary: the target stage cannot observe realization loss, the realization stage receives no end-to-end feedback, and transport enters the objective of neither. DOW-KE therefore removes the anchor and optimizes the layerwise weight updates directly against the final editing objective. No intermediate target is left that can fail to be realized, and transport, cross-layer coupling, and within-batch interference all move inside the forward pass through which that objective is evaluated. Only the anchor's role as an intermediate target is removed: the mechanism by which edit keys and preservation statistics predetermine an update's response scope is retained in full and folded into the update parameterization, so preservation and editing no longer compete inside a per-layer least-squares problem.

\section{Method}
DOW-KE builds on a structural observation about the closed-form update: for both MEMIT and AlphaEdit, Eq.~\ref{eq:closed-form delta W} factors into
\begin{equation}
	\Delta \;=\; R\,B,\qquad R\in\mathbf{R}^{d_1\times u},\quad B\in\mathbf{R}^{u\times d_0}.
	\label{eq:two factors}
\end{equation}
The columns of $R=[\mathbf{r}_1,\dots,\mathbf{r}_u]$ are the residuals allocated from the anchor target, so the anchor influences the update only through $R$; $B$ is a solution operator built solely from the edit keys and the preservation statistics, namely $K_1^{\top}(\lambda C+K_1K_1^{\top})^{-1}$ for MEMIT and $K_1^{\top}P[(K_1K_1^{\top}+K_pK_p^{\top})P+I]^{-1}$ for AlphaEdit, with $P$ the null-space projection derived from $C$ and $K_p$ the edit keys of previous batches.

DOW-KE keeps $B$ fixed during optimization and frees $R$: at each edited layer, the allocated residual is replaced by a directly optimized variable $A^{l}\in\mathbf{R}^{d_1\times u}$, giving $\Delta^{l}=A^{l}B^{l}$. Here $B^{l}$ is obtained by projecting a standard ridge basis onto the input directions allowed by AlphaEdit. Let $U_{\tau}^{l}$ contain the leading principal directions of $C^{l}$ whose cumulative spectral energy reaches at least $\tau$, and set $P^{l}=I-U_{\tau}^{l}(U_{\tau}^{l})^{\top}$. The variables $\{A^{l}\}$ are jointly optimized against the final editing objective of the deployed model. For any $\mathbf{k}\in\operatorname{span}(U_{\tau}^{l})$, the construction above gives $B^{l}\mathbf{k}=\mathbf{0}$; hence $\Delta^{l}\mathbf{k}=\mathbf{0}$ for every $A^{l}$. Thus, suppression on the selected key subspace is built into the parameterization rather than learned during optimization. The threshold $\tau$ controls the spectral-energy coverage of this protected subspace, and Table~\ref{tab:output_projection_ablation} shows that varying it shifts the empirical trade-off between Specificity and Generalization.

Preservation is applied on the output side as well. With $D_0$ the second moment of unrelated-knowledge outputs at layer $l$, we build $P_{\mathrm{out}}^{l}$ from $D_0$ by the same energy rule, giving the deployed update
\begin{equation}
	\Delta^{l} \;=\; P_{\mathrm{out}}^{l}\,A^{l}\,B^{l} .
	\label{eq:our delta W}
\end{equation}
$B^{l}$ constrains which inputs the update responds to; $P_{\mathrm{out}}^{l}$ keeps its write directions out of the subspace occupied by unrelated-knowledge outputs.

A third restriction is positional and acts only in the backward pass. The objective scores the target object at the end of the sequence, so it can also be reduced by writing at positions before the subject, which under causal attention have not read the complete subject and cannot be fact-directed. Writing $s_i$ for the subject's final position in rewriting sequence $i$ and $\mathrm{sg}[\cdot]$ for the stop-gradient operator, the increment contributed at position $t$ is $P_{\mathrm{out}}^{l}A^{l}(B^{l}\mathbf{k}_t^{l})$ for $t\ge s_i$ and $\mathrm{sg}[P_{\mathrm{out}}^{l}A^{l}](B^{l}\mathbf{k}_t^{l})$ for $t<s_i$. The two expressions are numerically identical, so the forward logits and loss are unchanged relative to the deployed update. Only the backward gradient routing changes: for positions before the subject, the direct gradient path through $P_{\mathrm{out}}^{l}A^{l}$ is stopped, while the path through $\mathbf{k}_t^{l}$ remains active, allowing the loss to backpropagate through the network.

Removing residual allocation also leaves the division of labor across layers to the end-to-end objective, so unfavorable transport at one layer can be absorbed by others; this is a by-product of joint optimization that we do not validate separately.

Collecting the pieces, DOW-KE solves
\begin{equation}
	\{A^{l}\}^{\star}=\arg\min_{\{A^{l}\}}\;
	\mathcal{L}_{\mathrm{edit}}\big(G_{\Delta}\big)
	+\lambda_{\mathrm{KL}}\,\mathcal{L}_{\mathrm{KL}}\big(G_{\Delta},G\big),
	\label{eq:dowke objective}
\end{equation}
where $G_{\Delta}$ is the model with the updates in Eq.~\ref{eq:our delta W} applied at all edited layers. We use the same losses as in Eq.~\ref{eq:anchor loss}: $\mathcal{L}_{\mathrm{edit}}$ is the negative log-likelihood of the target object, averaged over the $u$ requests and the $N$ prefixes $x_j$, while $\mathcal{L}_{\mathrm{KL}}$ is computed on the essence prompts $p_i'$. Unlike anchor optimization, both losses are evaluated on the model containing all layer updates and are used to optimize $\{A^{l}\}$ jointly. The positional rule is applied only when computing $\mathcal{L}_{\mathrm{edit}}$: it stops direct gradients to $A^{l}$ from positions before the subject without changing the forward outputs or the deployed updates. The KL loss uses the standard gradient path.

\section{Experiments}
\subsection{Models and Datasets}
To enable fair comparisons with prior work, we evaluate our method on three autoregressive language models commonly used in model-editing research: Llama-3-8B-Instruct \cite{metaai2024llama3}, GPT-J-6B \cite{wang2021gptj}, and GPT2-XL-1.5B \cite{radford2019language}. Following the standard configuration of locate-then-edit methods, all edits are applied to the FFN down-projection matrix at each layer within the layer window identified by causal tracing: {4–8} for Llama-3-8B-Instruct, {3–8} for GPT-J-6B, and {13–17} for GPT2-XL-1.5B \cite{fang2025alphaedit}. 
All experiments are run on NVIDIA L40 GPUs. We conduct experiments on CounterFact \cite{meng2022rome} and zsRE \cite{levy2017zeroshot}, two of the most influential benchmarks for model editing.

\subsection{Evaluation Metrics and Experiment Setup}
Efficacy, Generalization, and Specificity are the three core metrics for knowledge editing. They are measured on edit prompts, rephrased prompts, and neighborhood prompts, respectively, although their precise computation differ slightly between the two datasets. The main tables additionally report their harmonic mean as the overall Score. CounterFact further reports Fluency and Consistency. Fluency uses the n-gram entropy of generated text to detect repetitive degeneration, whereas Consistency measures the TF-IDF similarity between generated text and a reference passage describing the new fact, thereby assessing whether the edit is reflected in free-form generation. We follow the original AlphaEdit implementation for all metric calculations \cite{fang2025alphaedit}. Under the sequential editing protocol, we sample 2,000 requests from each dataset and apply them successively to the same model in 20 batches of 100 requests. Further details of the experimental setup are provided in Appendix A of the supplementary material.

\begin{table*}[t]
	\centering
	\footnotesize
	\setlength{\tabcolsep}{6pt}
	\renewcommand{\arraystretch}{1.0}
	\begin{tabular}{cl|cccccc|cccc}
		\hline
		\multirow{2}{*}{MODEL}
		& & \multicolumn{6}{c|}{CounterFact}
		& \multicolumn{4}{c}{zsRE} \\
		\cline{3-8}\cline{9-12}
		& Method & Score & Eff. & Gen. & Spe. & Flu. & Consis.
		&Score & Eff. & Gen. & Spe. \\
		\hline
		\multirow{8}{*}{LLaMA3-8B}
		& FT        & $62.24$ & $83.33$ & $67.79$ & $46.63$ & $233.72$ & $8.770$ & $23.00$ & $30.48$ & $30.22$ & $15.49$ \\
		& ROME      & $57.65$ & $64.40$ & $61.42$ & $49.44$ & $449.06$ & $3.31$ & $1.20$ & $2.01$ & $1.80$ & $0.69$ \\
		& MEMIT     & $59.89$ & $65.65$ & $64.65$ & $51.56$ & $437.43$ & $6.58$ & $26.10$ & $34.62$& $31.28$ & $18.49$\\
		& AlphaEdit & $84.61$ & $98.90$& $94.22$ & $67.88$ & $622.49$ & \underline{$32.40$} & \underline{$57.38$} & $94.47$& $91.13$ & \underline{$32.55$} \\
		& BLUE      & \underline{$89.34$} & $\bm{99.93}$ & $\bm{97.25}$ & \underline{$75.24$} & \underline{$624.90$} & $\bm{33.79}$ & $57.00$ & \underline{$95.77$} & \underline{$91.73$} & $31.96$ \\
		& FE*        & $84.95$& $98.35$ & $94.92$ & $68.43$ & - & - & $56.58$ & $92.96$ & $88.81$ & $32.25$  \\
		\hhline{~~|*{6}{-}|*{4}{-}}
		\rowcolor{highlight}
		\cellcolor{white} & \textbf{DOW-KE }   & $\bm{90.79}$ & \underline{$99.85$} & \underline{$96.64$} & $\bm{78.86}$ & $\bm{628.49}$& $31.63$ & $\bm{60.26}$ & $\bm{98.62}$ & $\bm{96.76}$ & $\bm{34.12}$ \\
		\hline
		\hline
		\multirow{8}{*}{GPT-J}
		& FT        & $62.85$ & $92.15$ & $72.38$ & $43.35$ & $297.92$ & $6.65$ & $37.88$ & $72.37$ & $68.91$ & $19.66$ \\
		& ROME      & $54.49$ & $57.50$ & $54.20$ & $52.05$ & $589.42$ & $3.22$ & $21.83$ & $56.42$ & $54.65$ & $9.86$ \\
		& MEMIT     & $82.57$ & $98.55$ & $95.50$ & $63.64$ & $546.28$ & $34.89$ & $55.02$ & $94.91$ & $90.22$ & \underline{$30.39$} \\
		& AlphaEdit & $89.23$ & $99.70$ & $96.68$ & \underline{$75.48$} & $618.50$ & $\bm{42.08}$ & $53.78$ & \underline{$99.79$} & \underline{$96.00$} &  $28.29$ \\
		& BLUE      & \underline{$89.26$} & \underline{$99.77$} & \underline{$97.13$} & $75.23$ & $\bm{621.07}$ &  \underline{$41.34$} & \underline{$54.21$} & $99.63$ & $95.96$ & $28.67$ \\
		& FE*        & $80.70$ & $97.4$ & $88.45$ & $64.1$ & - & - & $52.66$ & $98.62$ & $90.81$ & $27.92$ \\
		\hhline{~~|*{6}{-}|*{4}{-}}
		\rowcolor{highlight}
		\cellcolor{white}  & \textbf{DOW-KE}    & $\bm{90.93}$ & $\bm{99.90}$ & $\bm{98.48}$ & $\bm{77.96}$ & \underline{$619.02$} & $39.79$ & $\bm{61.09}$ & $\bm{99.81}$ & $\bm{98.50}$ & $\bm{34.56}$ \\
		\hline
		\hline
		\multirow{8}{*}{GPT2-XL}
		& FT        & $52.67$ & $63.55$ & $42.20$ & $57.06$ & $519.35$ & $10.56$ & $19.54$ & $37.11$ & $33.30$ & $10.36$ \\
		& ROME      & $52.78$ & $54.60$ & $51.18$ & $52.68$ & $366.13$ & $0.72$ & $26.30$ & $47.50$ & $43.56$ & $14.27$ \\
		& MEMIT     & $77.44$ & $94.70$ & $85.82$ & $60.50$ & $477.26$ & $22.72$ & $46.53$ & $79.17$ & $71.44$ & \underline{$26.42$} \\
		& AlphaEdit & $83.90$ & \underline{$99.50$} & $93.95$ & $66.39$ & $597.88$ & $39.38$ & $49.34$ & $94.81$ & $86.11$ & $25.88$ \\
		& BLUE      & $\bm{89.48}$ & $99.40$ & \underline{$96.00$} & $\bm{76.63}$ & \underline{$621.92$} & \underline{$40.98$} & \underline{$49.96$} & \underline{$96.88$} & \underline{$89.58$} & $25.93$ \\
		& FE*        & $77.96$ & $96.05$ & $88.75$ & $59.51$ & - & - & $34.66$ & $73.48$ & $64.11$ & $17.44$ \\
		\hhline{~~|*{6}{-}|*{4}{-}}
		\rowcolor{highlight}
		\cellcolor{white} & \textbf{DOW-KE}    & \underline{$87.66$} & $\bm{99.75}$ & $\bm{97.25}$ & \underline{$71.86$} & $\bm{625.93}$ & $\bm{41.65}$ & $\bm{54.44}$ & $\bm{98.25}$ & $\bm{95.27}$ & $\bm{29.04}$ \\
		\hline
	\end{tabular}
	\caption{Comparison of DOW-KE with existing methods on the sequential model editing task. The best results are highlighted in bold, while the second-best results are underlined. Baseline results are taken from \citet{fang2025alphaedit} and \citet{li2025rethinking}. For FE, no official results have been reported for the setting of 2,000 requests applied sequentially in 20 batches; the values in the table are from our own evaluation. Confidence intervals are reported in appendix of the supplementary material.}
	\label{table1}
\end{table*}

\begin{table}[t]
	\centering
	\small
	\setlength{\tabcolsep}{5pt}
	\renewcommand{\arraystretch}{1.0}
	
	\begin{tabular}{@{}llrrr@{}}
		\toprule
		\multicolumn{1}{c}{\multirow{2}{*}{MODEL}}
		& \multicolumn{4}{c}{CounterFact} \\
		\cmidrule{2-5}
		& Method
		& \multicolumn{1}{c}{Eff.}
		& \multicolumn{1}{c}{Gen.}
		& \multicolumn{1}{c}{Spe.} \\
		\midrule
		\multirow{3}{*}{LLaMA3-8B}
		& Baseline      & 98.3  & 90.05 & 77.59 \\
		& NLL-to-W.($\alpha=1.2$) & 99.25  & 95.58  & 77.85 \\
		&  + In-Graph C. & 99.85 & 95.62 & 80.24 \\
		&  + In-Graph C. + Routing & 99.85 & 96.64 & 78.86 \\
		\bottomrule
	\end{tabular}
	\caption{Operating-point-matched ablation results. The scaling coefficient $\alpha$ controls the trade-off between editing performance and Specificity. Rows 1--2 compare Efficacy and Generalization at approximately matched Specificity, whereas Rows 2--3 compare Specificity at approximately matched editing performance. }
	\label{table2}
\end{table}
\subsection{Comparative Experiments}
Table \ref{table1} reports DOW-KE's evaluation results on three models and two datasets. All values are averaged over five runs; under the same experimental settings, we compare DOW-KE with other locate-then-edit methods and fine-tuning-based baselines \cite{fang2025alphaedit}. DOW-KE achieves the best result in most settings and the second-best result in the remaining settings. Compared with BLUE, it has a clear advantage in five of the six model–dataset combinations; in particular, it outperforms BLUE on every reported metric on zsRE. We also include FE as a comparison method, whose results are relatively weak under the multi-batch sequential-editing protocol. Thus, DOW-KE not only consistently improves on AlphaEdit but also outperforms, overall, methods that retain the anchor structure, including BLUE and FE.

\subsection{Ablation Studies}
Table \ref{table2} examines two key design choices of DOW-KE: removing the intermediate activation anchor and directly optimizing the weights, and incorporating the preservation-subspace constraint into the differentiable computation graph. We use AlphaEdit with the energy-threshold rule as the comparison baseline. Although this setting differs slightly from the original AlphaEdit, it uses the same thresholding rule as DOW-KE. Let $\alpha$ denote the weight-scaling coefficient, $\Delta W = \alpha \Delta W_{\mathrm{grad}}$, with a default value of 1.  
We treat $\alpha$ as an operating-point control along the trade-off between editing performance and Specificity. At matched Specificity (77.59 versus 77.85), NLL-to-W. with $\alpha=1.2$ improves Efficacy and Generalization over the baseline, demonstrating the benefit of direct weight optimization. At comparable Efficacy and Generalization, adding the in-graph constraint with the default $\alpha=1$ improves Specificity from 77.85 to 80.24. These matched comparisons support the complementary contributions of direct optimization and in-graph constraint parameterization.
Row 4 adds subject-conditioned gradient routing, which is designed to prevent the optimizer from directly shaping $A^l$ through positions that precede and therefore do not fully encode the subject. Routing improves Generalization by 1.02 points while reducing Specificity by 1.38 points, resulting in a modest 0.31-point decrease in Score. We include it in the full configuration for its structural role in promoting fact-conditioned optimization, rather than as a score-maximizing component.

\begin{table}[t]
	\centering
	\small
	\setlength{\tabcolsep}{6pt}
	\renewcommand{\arraystretch}{1.0}

	\begin{tabular*}{\columnwidth}{@{\extracolsep{\fill}}clrrr@{}}
		\toprule
		\multirow{2}{*}{Threshold}
		& \multirow{2}{*}{Out. Proj.}
		& \multicolumn{3}{c}{CounterFact} \\
		\cmidrule{3-5}
		&
		& \multicolumn{1}{c}{Eff.}
		& \multicolumn{1}{c}{Gen.}
		& \multicolumn{1}{c}{Spe.} \\
		\midrule

		\multirow{2}{*}{$\tau=0.4$}
		& w.   & 99.94 & 97.45 & 71.59 \\
		& w/o. & 99.95 & 97.05 & 71.02 \\
		\cmidrule{1-5}

		\multirow{2}{*}{$\tau=0.5$}
		& w.   & 99.93 & 97.12 & 76.98 \\
		& w/o. & 99.95 & 96.37 & 74.62 \\
		\cmidrule{1-5}

		\multirow{2}{*}{$\tau=0.6$}
		& w.   & 99.85 & 95.62 & 80.24 \\
		& w/o. & 99.91 & 96.12 & 79.65 \\
		\cmidrule{1-5}

		\multirow{2}{*}{$\tau=0.7$}
		& w.   & 99.86 & 94.74 & 84.63 \\
		& w/o. & 99.85 & 95.33 & 83.22 \\
		\bottomrule
	\end{tabular*}
	\caption{Ablation results for the subspace threshold and output-side projection on LLaMA3-8B and CounterFact. Experiments are conducted without subject routing to isolate the effects of the subspace threshold and output-side projection.}
	\label{tab:output_projection_ablation}
\end{table}

Table~\ref{tab:output_projection_ablation} shows a threshold-controlled trade-off: increasing $\tau$ improves Specificity but reduces Generalization. One possible explanation is that the enlarged preservation subspace leaves fewer editable directions, encouraging narrower fitting to the optimization prompts and weaker transfer to paraphrases. We leave direct verification of this mechanism to future work. Table \ref{tab:output_projection_ablation} also compares performance with and without the output-side preservation constraint. Across all thresholds shown, adding output-side preservation improves Specificity and, in some settings, also improves Generalization. This suggests that output-side preservation provides a constraint complementary to key-side preservation, thereby strengthening the protection of unrelated knowledge.

%

\section{Conclusion}
We introduce DOW-KE, an anchor-free knowledge-editing method that addresses the substantial gap between pre-specified intermediate targets and their actual realization in multi-layer editing. DOW-KE removes the intermediate activation anchor and layerwise residual allocation, and brings preservation constraints, transport attenuation, directional distortion, and cross-layer coupling into a single differentiable computation graph. The update that is optimized is therefore exactly the update that is deployed, eliminating the mechanism by which a prescribed intermediate target can remain unrealized. At the same time, the method retains the mechanism by which edit keys and preservation statistics predetermine the response scope of an update, so that each edit remains tied to a specific fact. Experiments on three models and two datasets show that DOW-KE achieves the highest composite Score in five of six evaluated settings.

	
	\bibliography{references}

@inproceedings{decao2021editing,
  author    = {De Cao, Nicola and Aziz, Wilker and Titov, Ivan},
  title     = {Editing Factual Knowledge in Language Models},
  editor    = {Moens, Marie-Francine and Huang, Xuanjing and Specia, Lucia and Yih, Scott Wen-tau},
  booktitle = {Proceedings of the 2021 Conference on Empirical Methods in Natural Language Processing},
  year      = {2021},
  month     = nov,
  address   = {Online and Punta Cana, Dominican Republic},
  pages     = {6491--6506},
  publisher = {Association for Computational Linguistics},
  doi       = {10.18653/v1/2021.emnlp-main.522},
  url       = {https://aclanthology.org/2021.emnlp-main.522/}
}

@inproceedings{hartvigsen2023grace,
  author    = {Hartvigsen, Tom and Sankaranarayanan, Swami and Palangi, Hamid and Kim, Yoon and Ghassemi, Marzyeh},
  title     = {Aging with {GRACE}: Lifelong Model Editing with Discrete Key-Value Adaptors},
  editor    = {Oh, A. and Naumann, T. and Globerson, A. and Saenko, K. and Hardt, M. and Levine, S.},
  booktitle = {Advances in Neural Information Processing Systems},
  year      = {2023},
  volume    = {36},
  pages     = {47934--47959},
  publisher = {Curran Associates, Inc.},
  doi       = {10.52202/075280-2079},
  url       = {https://proceedings.neurips.cc/paper_files/paper/2023/file/95b6e2ff961580e03c0a662a63a71812-Paper-Conference.pdf}
}

@inproceedings{jiang2024instruction,
  author    = {Jiang, Zhengbao and Sun, Zhiqing and Shi, Weijia and Rodriguez, Pedro and Zhou, Chunting and Neubig, Graham and Lin, Xi Victoria and Yih, Wen-tau and Iyer, Srinivasan},
  title     = {Instruction-tuned Language Models are Better Knowledge Learners},
  editor    = {Ku, Lun-Wei and Martins, Andre and Srikumar, Vivek},
  booktitle = {Proceedings of the 62nd Annual Meeting of the Association for Computational Linguistics (Volume 1: Long Papers)},
  year      = {2024},
  month     = aug,
  address   = {Bangkok, Thailand},
  pages     = {5421--5434},
  publisher = {Association for Computational Linguistics},
  doi       = {10.18653/v1/2024.acl-long.296},
  url       = {https://aclanthology.org/2024.acl-long.296/}
}

@inproceedings{li2025rethinking,
  author    = {Li, Xiaopeng and Wang, Shangwen and Li, Shasha and Song, Shezheng and Ji, Bin and Ma, Jun and Yu, Jie},
  title     = {Rethinking Residual Distribution in Locate-then-Edit Model Editing},
  editor    = {Belgrave, D. and Zhang, C. and Lin, H. and Pascanu, R. and Koniusz, P. and Ghassemi, M. and Chen, N.},
  booktitle = {Advances in Neural Information Processing Systems},
  year      = {2025},
  volume    = {38},
  pages     = {78348--78374},
  publisher = {Curran Associates, Inc.},
  url       = {https://proceedings.neurips.cc/paper_files/paper/2025/file/70d4ef44dc973586cfa3ea92b4868b72-Paper-Conference.pdf}
}

@inproceedings{liu2026forwardreplay,
  author    = {Liu, Wei and Liu, Hongkai and Deng, Zhiying and Teh, Yee Whye and Lee, Wee Sun},
  title     = {From Backward Spreading to Forward Replay: Revisiting Target Construction in {LLM} Parameter Editing},
  booktitle = {Proceedings of the 43rd International Conference on Machine Learning},
  year      = {2026},
  address   = {Seoul, South Korea},
  series    = {Proceedings of Machine Learning Research},
  publisher = {PMLR},
  url       = {https://icml.cc/virtual/2026/poster/61038}
}

@inproceedings{meng2022rome,
  author    = {Meng, Kevin and Bau, David and Andonian, Alex and Belinkov, Yonatan},
  title     = {Locating and Editing Factual Associations in {GPT}},
  editor    = {Koyejo, S. and Mohamed, S. and Agarwal, A. and Belgrave, D. and Cho, K. and Oh, A.},
  booktitle = {Advances in Neural Information Processing Systems},
  year      = {2022},
  volume    = {35},
  pages     = {17359--17372},
  publisher = {Curran Associates, Inc.},
  doi       = {10.52202/068431-1262},
  url       = {https://proceedings.neurips.cc/paper_files/paper/2022/file/6f1d43d5a82a37e89b0665b33bf3a182-Paper-Conference.pdf}
}

@inproceedings{meng2023memit,
  author    = {Meng, Kevin and Sharma, Arnab Sen and Andonian, Alex J. and Belinkov, Yonatan and Bau, David},
  title     = {Mass-Editing Memory in a Transformer},
  booktitle = {The Eleventh International Conference on Learning Representations},
  year      = {2023},
  address   = {Kigali, Rwanda},
  publisher = {OpenReview.net},
  url       = {https://openreview.net/forum?id=MkbcAHIYgyS}
}

@inproceedings{mitchell2022mend,
  author    = {Mitchell, Eric and Lin, Charles and Bosselut, Antoine and Finn, Chelsea and Manning, Christopher D.},
  title     = {Fast Model Editing at Scale},
  booktitle = {The Tenth International Conference on Learning Representations},
  year      = {2022},
  address   = {Virtual Event},
  publisher = {OpenReview.net},
  url       = {https://openreview.net/forum?id=0DcZxeWfOPt}
}

@misc{eskandar2026loki,
  author        = {Eskandar, Masih and Sirera Perell{\'o}, Miquel and Ioannidis, Stratis and Dy, Jennifer},
  title         = {{LOKI}: Memory-Free Null-Space Constrained Lifelong Knowledge Editing},
  year          = {2026},
  month         = jun,
  note          = {arXiv preprint},
  eprint        = {2606.19679},
  archiveprefix = {arXiv},
  primaryclass  = {cs.LG},
  doi           = {10.48550/arXiv.2606.19679},
  url           = {https://arxiv.org/abs/2606.19679}
}

@inproceedings{fang2025alphaedit,
  author    = {Fang, Junfeng and Jiang, Houcheng and Wang, Kun and Ma, Yunshan and Shi, Jie and Wang, Xiang and He, Xiangnan and Chua, Tat-Seng},
  title     = {{AlphaEdit}: Null-Space Constrained Knowledge Editing for Language Models},
  booktitle = {The Thirteenth International Conference on Learning Representations},
  year      = {2025},
  address   = {Singapore},
  publisher = {OpenReview.net},
  url       = {https://openreview.net/forum?id=HvSytvg3Jh}
}

@inproceedings{lyu2026evoedit,
  author    = {Lyu, Sicheng and Gu, Yu and Wang, Xinyu and Huang, Jerry and Luan, Sitao and Cui, Yufei and Chang, Xiao-Wen and Lu, Peng},
  title     = {{EvoEdit}: Evolving Null-space Alignment for Robust and Efficient Knowledge Editing},
  editor    = {Liakata, Maria and Moreira, Viviane P. and Zhang, Jiajun and Jurgens, David},
  booktitle = {Findings of the Association for Computational Linguistics: {ACL} 2026},
  year      = {2026},
  month     = jul,
  address   = {San Diego, California, United States},
  pages     = {1520--1540},
  publisher = {Association for Computational Linguistics},
  doi       = {10.18653/v1/2026.findings-acl.75},
  url       = {https://aclanthology.org/2026.findings-acl.75/}
}

@inproceedings{wang2024wise,
  author    = {Wang, Peng and Li, Zexi and Zhang, Ningyu and Xu, Ziwen and Yao, Yunzhi and Jiang, Yong and Xie, Pengjun and Huang, Fei and Chen, Huajun},
  title     = {{WISE}: Rethinking the Knowledge Memory for Lifelong Model Editing of Large Language Models},
  editor    = {Globerson, A. and Mackey, L. and Belgrave, D. and Fan, A. and Paquet, U. and Tomczak, J. and Zhang, C.},
  booktitle = {Advances in Neural Information Processing Systems},
  year      = {2024},
  volume    = {37},
  address   = {Rio de Janeiro, Brazil},
  pages     = {53764--53797},
  publisher = {Curran Associates, Inc.},
  doi       = {10.52202/079017-1703},
  url       = {https://proceedings.neurips.cc/paper_files/paper/2024/hash/60960ad78868fce5c165295fbd895060-Abstract-Conference.html}
}

@inproceedings{yang2026finetuning,
  author    = {Yang, Wanli and Tang, Rui and Zang, Hongyu and Su, Du and Cao, Qi and Wang, Jingang and Shen, Huawei and Cheng, Xueqi and Sun, Fei},
  title     = {Fine-tuning Done Right in Model Editing},
  booktitle = {The Fourteenth International Conference on Learning Representations},
  year      = {2026},
  publisher = {OpenReview.net},
  url       = {https://openreview.net/forum?id=cfHuA5jsPt}
}

@article{yu2024melo,
  author  = {Yu, Lang and Chen, Qin and Zhou, Jie and He, Liang},
  title   = {{MELO}: Enhancing Model Editing with Neuron-Indexed Dynamic {LoRA}},
  journal = {Proceedings of the AAAI Conference on Artificial Intelligence},
  year    = {2024},
  volume  = {38},
  number  = {17},
  pages   = {19449--19457},
  doi     = {10.1609/aaai.v38i17.29916},
  url     = {https://ojs.aaai.org/index.php/AAAI/article/view/29916}
}

@inproceedings{levy2017zeroshot,
  author    = {Levy, Omer and Seo, Minjoon and Choi, Eunsol and Zettlemoyer, Luke},
  title     = {Zero-Shot Relation Extraction via Reading Comprehension},
  editor    = {Levy, Roger and Specia, Lucia},
  booktitle = {Proceedings of the 21st Conference on Computational Natural Language Learning ({CoNLL} 2017)},
  year      = {2017},
  month     = aug,
  address   = {Vancouver, Canada},
  pages     = {333--342},
  publisher = {Association for Computational Linguistics},
  doi       = {10.18653/v1/K17-1034},
  url       = {https://aclanthology.org/K17-1034/}
}

@misc{metaai2024llama3,
  author       = {{Meta AI}},
  title        = {Introducing {Meta Llama 3}: The Most Capable Openly Available {LLM} to Date},
  year         = {2024},
  month        = apr,
  howpublished = {\url{https://ai.meta.com/blog/meta-llama-3/}},
  note         = {Accessed: 2026-07-25}
}

@techreport{radford2019language,
  author      = {Radford, Alec and Wu, Jeffrey and Child, Rewon and Luan, David and Amodei, Dario and Sutskever, Ilya},
  title       = {Language Models are Unsupervised Multitask Learners},
  institution = {OpenAI},
  year        = {2019},
  month       = feb,
  type        = {Technical report},
  note        = {Available at \url{https://cdn.openai.com/better-language-models/language-models.pdf}}
}

@misc{wang2021gptj,
  author       = {Wang, Ben and Komatsuzaki, Aran},
  title        = {{GPT-J-6B}: A 6 Billion Parameter Autoregressive Language Model},
  year         = {2021},
  month        = may,
  howpublished = {\url{https://github.com/kingoflolz/mesh-transformer-jax}},
  note         = {EleutherAI model release; accessed: 2026-07-25}
}

@inproceedings{geva2023dissecting,
  author    = {Geva, Mor and Bastings, Jasmijn and Filippova, Katja and Globerson, Amir},
  title     = {Dissecting Recall of Factual Associations in Auto-Regressive Language Models},
  editor    = {Bouamor, Houda and Pino, Juan and Bali, Kalika},
  booktitle = {Proceedings of the 2023 Conference on Empirical Methods in Natural Language Processing},
  year      = {2023},
  month     = dec,
  address   = {Singapore},
  pages     = {12216--12235},
  publisher = {Association for Computational Linguistics},
  doi       = {10.18653/v1/2023.emnlp-main.751},
  url       = {https://aclanthology.org/2023.emnlp-main.751/}
}

@inproceedings{lewis2020retrieval,
  author    = {Lewis, Patrick and Perez, Ethan and Piktus, Aleksandra and Petroni, Fabio and Karpukhin, Vladimir and Goyal, Naman and K{\"u}ttler, Heinrich and Lewis, Mike and Yih, Wen-tau and Rockt{\"a}schel, Tim and Riedel, Sebastian and Kiela, Douwe},
  title     = {Retrieval-Augmented Generation for Knowledge-Intensive {NLP} Tasks},
  editor    = {Larochelle, Hugo and Ranzato, Marc'Aurelio and Hadsell, Raia and Balcan, Maria-Florina and Lin, Hsuan-Tien},
  booktitle = {Advances in Neural Information Processing Systems},
  year      = {2020},
  volume    = {33},
  pages     = {9459--9474},
  publisher = {Curran Associates, Inc.},
  url       = {https://proceedings.neurips.cc/paper/2020/hash/6b493230205f780e1bc26945df7481e5-Abstract.html}
}

@inproceedings{hase2023localization,
  author    = {Hase, Peter and Bansal, Mohit and Kim, Been and Ghandeharioun, Asma},
  title     = {Does Localization Inform Editing? Surprising Differences in Causality-Based Localization vs. Knowledge Editing in Language Models},
  editor    = {Oh, A. and Naumann, T. and Globerson, A. and Saenko, K. and Hardt, M. and Levine, S.},
  booktitle = {Advances in Neural Information Processing Systems},
  year      = {2023},
  volume    = {36},
  pages     = {17643--17668},
  publisher = {Curran Associates, Inc.},
  doi       = {10.52202/075280-0774},
  url       = {https://proceedings.neurips.cc/paper_files/paper/2023/hash/3927bbdcf0e8d1fa8aa23c26f358a281-Abstract-Conference.html}
}

@inproceedings{sinitsin2020editable,
  author    = {Sinitsin, Anton and Plokhotnyuk, Vsevolod and Pyrkin, Dmitry and Popov, Sergei and Babenko, Artem},
  title     = {Editable Neural Networks},
  booktitle = {The Eighth International Conference on Learning Representations},
  year      = {2020},
  publisher = {OpenReview.net},
  url       = {https://openreview.net/forum?id=HJedXaEtvS}
}

@inproceedings{yao2023editing,
  author    = {Yao, Yunzhi and Wang, Peng and Tian, Bozhong and Cheng, Siyuan and Li, Zhoubo and Deng, Shumin and Chen, Huajun and Zhang, Ningyu},
  title     = {Editing Large Language Models: Problems, Methods, and Opportunities},
  editor    = {Bouamor, Houda and Pino, Juan and Bali, Kalika},
  booktitle = {Proceedings of the 2023 Conference on Empirical Methods in Natural Language Processing},
  year      = {2023},
  month     = dec,
  address   = {Singapore},
  pages     = {10222--10240},
  publisher = {Association for Computational Linguistics},
  doi       = {10.18653/v1/2023.emnlp-main.632},
  url       = {https://aclanthology.org/2023.emnlp-main.632/}
}
	
\end{document}